# DerainCycleGAN: Rain Attentive CycleGAN for Single Image Deraining and Rainmaking

Yanyan Wei [1], Zhao Zhang [1], Yang Wang [1], Mingliang Xu [2], Yi Yang [3],
Shuicheng Yan [4], Meng Wang [1]

[1] Hefei University of Technology,  [2] Zhengzhou University, 3 University of Technology Sydney
[4] National University of Singapore

**Abstract**

*Single image deraining (SID) is an important and challenging topic in emerging vision applications, and most of emerged deraining methods are supervised relying on the ground truth (i.e., paired images) in recent years. However, in practice it is rather common to have no unpaired images in real deraining task, in such cases how to remove the rain streaks in an unsupervised way will be a very challenging task due to lack of constraints between images and hence suffering from low-quality recovery results. In this paper, we explore the unsupervised SID task using unpaired data and propose a novel net called Attention-guided Deraining by Constrained CycleGAN (or shortly, DerainCycleGAN), which can fully utilize the constrained transfer learning abilitiy and circulatory structure of CycleGAN. Specifically, we design an unsupervised attention guided rain streak extractor (U-ARSE) that utilizes a memory to extract the rain streak masks with two constrained cycle-consistency branches jointly by paying attention to both the rainy and rain-free image domains. As a by-product, we also contribute a new paired rain image dataset called Rain200A, which is constructed by our network automatically. Compared with existing synthesis datasets, the rainy streaks in Rain200A contains more obvious and diverse shapes and directions. As a result, existing supervised methods trained on Rain200A can perform much better for processing real rainy images. Extensive experiments on synthesis and real datasets show that our net is superior to existing unsupervised deraining networks, and is also very competitive to other related supervised networks.*

## 1. Introduction

Images and videos captured in rainy days from outdoor vision system, e.g., self-driving, surveillance and person/vehicle tracking, are usually degenerated by the rain streaks and drops. This will decrease subsequent high-level tasks, for instance object detection [2], image recognition [5] and saliency detection [9], directly. As such, image deraining, especially for the single image deraining (SID), has a wide range of applications. However, due to the irregular complex rain streaks in practice and ill-posed property, SID still remains an important but challenging unmanageable issue so far. SID considers removing the rain streak component $R$ and recovering the clean background image $B$ from rainy image $X$, which can be formulated as follows:

$$X = R + B.  \qquad (1)$$

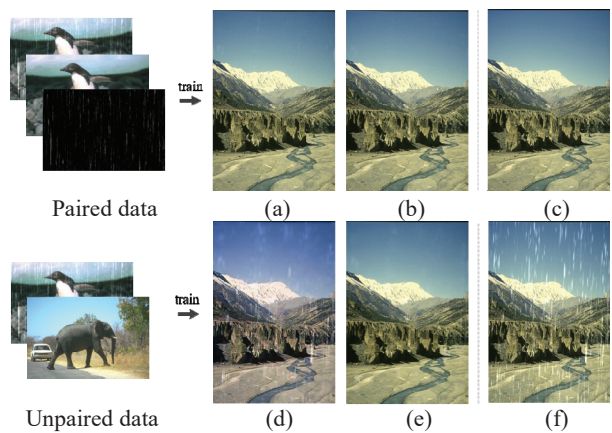

Figure 1. Comparison of supervised and unsupervised deraining on Rain100L [15], where (a) and (b) are the results of supervised JORDER [15] and PReNet [24], respectively; (d) and (e) are the unsupervised deraining results of CycleGAN [1] and our network, respectively; (c) and (f) denote the ground truth and original rainy images, respectively. The results show that our net obtains competitive results to the supervised nets. Moreover, our net can keep the color and structure of images, compared with CycleGAN [1].

To solve Eqn.(1), lots of deep learning based deraining networks [3, 4, 12, 13, 14, 15, 16, 22, 24, 25, 26, 27, 28, 40] have been proposed in recent years. It is noteworthy that most existing methods are supervised ones that are trained in a supervised way on the synthetic datasets [4, 10, 13, 15], as they explicitly require the pairs of rainy and clean images (ground-truth). By defining a strict constraint between the rainy image and its ground-truth, supervised models can usually obtain the promising deraining results (see Figs.1(a) and (b)). The enhanced performance can be attributed to the fact that the rain streaks of rainy images have been trained already, i.e., the rain streaks of input rain images are in fact already known to the models. As such, supervised methods usually have a strong generalization ability using the paired data. However, for real rainy images without ground-truth (i.e., unpaired), most existing supervised deraining models may fail due to the irregularity and nonuniform rain steaks, which can be seen from the example in Fig.2.

It is noteworthy that most of rainy images captured from the real world have no ground-truth, so they cannot be used directly by the supervised nets. In such cases, unsupervised or semi-supervised methods will be descried, since they can perform deraining without needing paired images or with a just small amount of paired images. Due to the lack of prior knowledge, the researches on semi-supervised and unsupervised nets for the task of SID develop much slower than supervised ones. The main reasons are twofold: (1) the rain streaks of real rainy images have very irregular shapes and directions (e.g., streak, drop and veil). Even for the synthesis datasets, it is still difficult to learn an accurate mapping between the rainy and rain-free images without certain strict pairwise constraint; (2) for existing synthesis datasets (e.g., Rain1400 [4] and Rain800 [13]), the fully-supervised methods still cannot obtain ideal recovery results, i.e., there is still a lot of space for improvement.

In this work, we design an unsupervised SID network by unpaired images. The main contributions are as follows:

(1) A novel unsupervised SID framework called Attention-guided Deraining by Constrained CycleGAN (shortly, DerainCycleGAN) is proposed. DerainCycleGAN directly performs the task of SID on unpaired images, i.e., without using ground-truth. To the best of our knowledge, this is one of few unsupervised nets of using CycleGAN for SID. Specifically, our network first utilizes the attention guided transfer ability between rainy and rain-free images, and the circulatory structures of CycleGAN with two constrained branches for SID. Extensive results on several challenging synthetic and real rainy image datasets illustrate that our net can deliver competitive results to existing semi-supervised, unsupervised and even supervised methods.

(2) To extract the rain streaks from rainy images accurately, a new and unsupervised attention guided rain streak extractor (U-ARSE) is presented, which can learn the rain streak masks using unpaired images. Specifically, U-ARSE pays attention to both rain and rain-free image domains and can approximately extract the rain streaks stage by stage, as shown in Fig.4. This operation can also resolve the information asymmetry in the two domains. Due to lack of the paired information and strict constraint, we equip U-ARSE with multi-losses, which enables it to extract more clear and precise rain streaks and make rain at the same time.

(3) A new paired rain image dataset called Rain200A is automatically created as a by-product from the rain-free to rain-free cycle-consistency branch of our net, as shown in

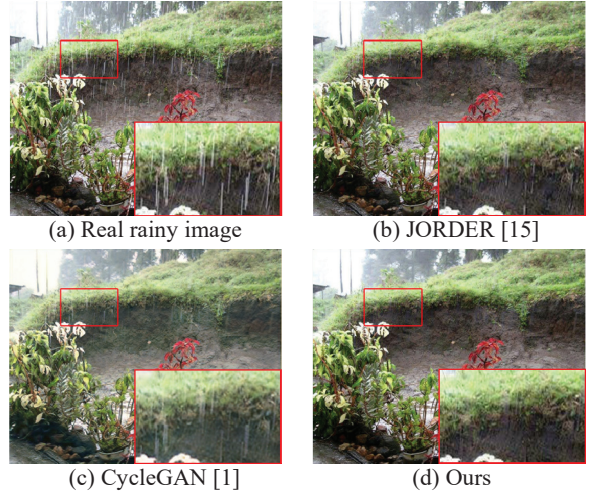

(a) Real rainy image      (b) JORDER [15]

(c) CycleGAN [1]      (d) Ours

Figure 2. Illustration of the recovery results on real images, where (a) is the original real rainy image without ground-truth; (b) is the result of JORDER trained in an supervised way on Rain100L (i.e., with paired images); (c) and (d) are the results of CycleGAN and our net trained without paired images in an unsupervised way on Rain100L, respectively. We find JORDER and CycleGAN cannot effectively remove the rain streaks and CycleGAN even tends to blur it, while our net performs well in recovering the background.

Fig.3. To the best of our knowledge, this is the first dataset for SID with the generated irregular rain steaks and shapes. Due to the transfer learning ability and circulatory structure, Rain200A contains much more types of rain streaks and is more similar to those of real rain images than most existing synthesis rain datasets that usually contains well-designed rain streaks. The results in Fig.5 clearly show that existing supervised deraining net trained on Rain200A can deliver better performance on real rainy images than those trained on existing synthesis rain image datasets.

## 2. Related Work

### 2.1. GAN and CycleGAN

Generative Adversarial Networks (GAN) [29] are deep neural net architectures including two sub-nets, pitting one against the other (i.e., "adversarial"). GAN is effective in generating more realistic images. However, most existing GAN based models require the paired training data, which is usually expensive to obtain in practice. To address this issue, an unsupervised GAN, termed CycleGAN [1], has been recently proposed using unpaired images. CycleGAN translates an image from a source domain $X$ to a target domain $Y$ in the absence of paired data. Specifically, CycleGAN learns a mapping $G: X \rightarrow Y$ such that the distribution of images from $G(X)$ is indistinguishable from $Y$ by an adversarial loss. However, because the mapping $G: X \rightarrow Y$ is highly under-constrained, CycleGAN couples it with an inverse mapping $F: Y \rightarrow X$ and introduces a cycle consistency loss to enforce $F(G(X)) \approx X$ (and vice versa). Recently, the CycleGAN-based methods have obtained great access in various vision tasks, e.g., image deblurring [31], image dehazing [34] and image super-resolution [33].

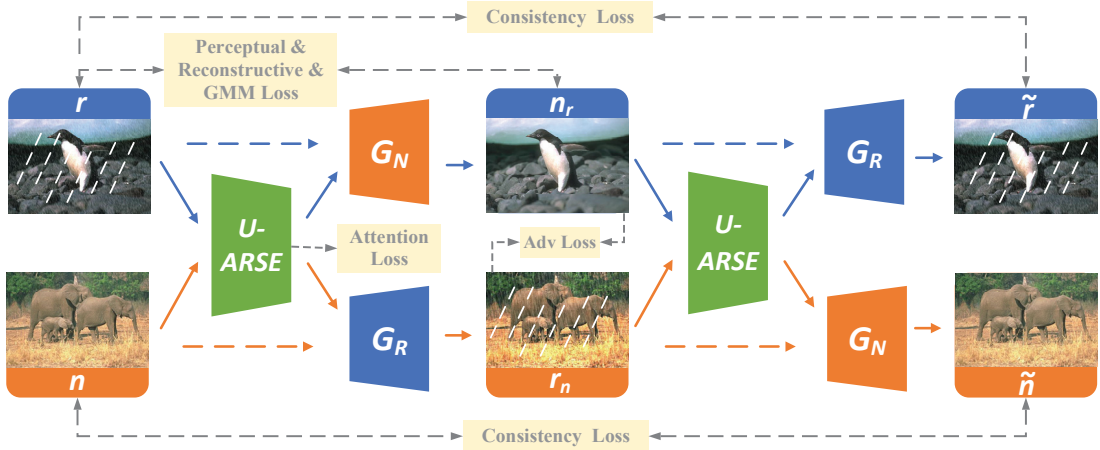

Figure 3. The pipeline of our detraining net that has three parts: (1) An unsupervised attention-based rain streak extractor (U-ARSE) that pays attention to the rain streak information in both rainy and rain free image domains; (2) Two generators $G_N$ and $G_R$ that can generate rain-free and rainy images respectively; (3) Two discriminators $D_N$ and $D_R$ that can distinguish the real image from faked image produced by generators. This network also has two constrained branches: (1) rainy to rainy cycle-consistency branch $r \rightarrow n_r \rightarrow \tilde{r}$, where the rainy image is used to generate rain-free image and is then reconstructed again by the generator. (2) rain-free to rain-free cycle-consistency branch $n \rightarrow r_n \rightarrow \tilde{n}$, where rain-free image is used to generate rainy image and is then reconstructed by the generator.

**Remarks.** Although CycleGAN has been proved to be effective in various low-level tasks, it is still hard to use CycleGAN to solve the rain removal problem in Eqn. (1) due to the fact that the domain knowledge of the rainy and rain-free images is asymmetrical. Specifically, rainy image contains the background and rain streaks (or drops), but the rain-free image only has background. As a result, directly utilizing CycleGAN for the task of SID may suffer from the color-/structure-destroying issue (see Fig.1(d)) and moreover it cannot fully recover the image blurs, colors and rain streaks. Although we used the identity loss (*identity loss = 0.1*), mentioned in the training step of CycleGAN for preserving the color of generated images. Under this circumstance, we propose a new network based on the CycleGAN, which can recurrently extract the rain streaks from the rainy images. It is worth noting that our network is fundamentally different from CycleGAN in following aspects. (1) our network is particularly designed for the SID, so the color and structure of images can be well preserved (see Fig.1(e)) by the multi-loss constrained unsupervised attention guided rain streak extractor (U-ARSE); (2) we make full use of the circulatory structures of CycleGAN by equipping it with two constrained branches for the SID. Specifically, we use the inverse mapping $n \rightarrow r_n \rightarrow \tilde{n}$ in the second branch to create a new rain image dataset Rain200A that contains 200 pairs of rain-free and rainy images from a new perspective.

## 2.2. GAN-based Deep Networks for SID

Some researches of using GAN to solve the rain removal problem have been recently proposed, such as [22, 28, 40]. Qian *et al.* [28] proposed a network to focus on solving the rain drop removal task using a generator and a discriminator. Zhang *et al.* [40] proposed a GAN-based deraining network by considering the quantitative, visual and also discriminative properties to define the problem using three losses. These two methods are supervised models with only one generator and one discriminator. More recently, Zhu *et al.* [22] proposed a GAN based unsupervised end-to-end adversarial deraining net, RainRemoval-GAN (RR-GAN), which can generate the realistic rain-free images using only the unpaired images. RR-GAN mainly defines a new multiscale attention memory generator and a novel multiscale deeply supervised discriminator, so it performs similarly as the above supervised GAN-based methods. As result, by only considering the consistency loss directly by adding the derained images to the extracted rain streaks, the constraint between the rainy and rain-free images is weak and even ill-posed. Besides, RR-GAN only trains an ARSE by using rainy images, and does not utilize important information in rain-free images, so it cannot extract the precise rain streaks from rainy images due to the ill-posed property of Eqn.(1). In contrast, our network takes full advantage of information in both rainy and rain-free image domains and constructs two pairs of generators and discriminators to construct a two-branch network, which can provide a more stable and reliable constraint. As described in Fig.7 and Table 1, we find that our net performs better than RR-GAN [22].

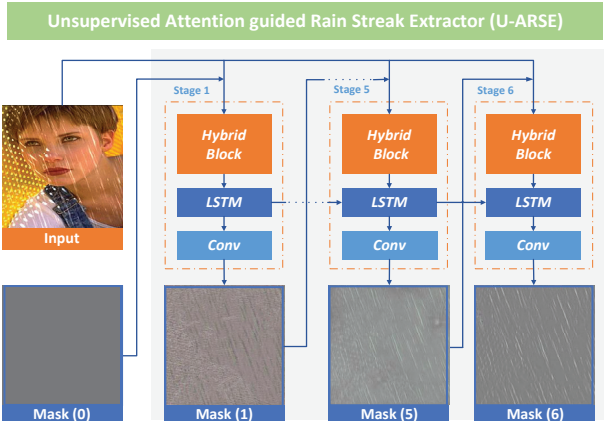

Figure 4. The structure of U-ARSE, where input image is sent to U-ARSE stage by stage. It can be seen that the mask is recurrently improved by the contexts from the previous stages while noise and background are also disappeared. The final rain streak mask (6) is very similar to the rain streak in input image, which can prove that our U-ARSE is effective. The input image together with the final are used as the input to generate the recovered image.

## 3. Proposed DerainCycleGAN

### 3.1. Network Architecture

The network pipeline of our DerainCycleGAN is shown in Fig. 3, which has three parts: (1) U-ARSE that extracts the rain streak masks from the rainy images stage by stage; (2) two generators $G_N$ and $G_R$ that can generate the rain-free and rainy images respectively; and (3) two discriminators $D_N$ and $D_R$ that can distinguish the real images from faked images obtained by the generators. DerainCycleGAN also includes two branches: (1) rainy to rainy cycle-consistency branch $r \to n_r \to \tilde{r}$, where rainy image is used to generate rain-free image and then reconstructed again by generator; and (2) rain-free to rain-free cycle-consistency branch $n \to r_n \to \tilde{n}$, where the rain-free image is used to generate rainy image and then reconstructed by generator.

### 3.2. Unsupervised Attention guided Rain Streak Extractor (U-ARSE)

The visual attention focuses on the important regions in images to capture the characteristics of that region. This idea has been proved to be efficient for the supervised SID [24, 28] and unsupervised SID [22], since it can make the networks know where the attention should be focused and make the SID task more precisely. But ARSE in [22, 24, 28] is essentially a single task module to extract the rain streaks, which only considers a mapping $G: X \to Y$ in a single branch. In supervised mode, ARSE extracts the precise rain streaks using supervised constraint, but this constraint is weak and is not stable in unsupervised mode. As such, we present U-ARSE to discover and focus on the rain streaks in two constrained cycle-consistency branches, that is, our U-ARSE pays attention to the rainy and rain-free images at the same time. Thus, the extracted rain streaks will be more accurate than ARSE that only focuses on the rainy images.

Technically, our U-ARSE module has 6 stages, as shown in Fig.4, where each stage includes a *Hybrid Block* unit (i.e., dual-path residual dense block [26]), a *LSTM* unit [35] and a *Convolutional layer*. The *Hybrid Block* has two paths (i.e., Residual path and Dense path) that can reuse the common feature from previous layers and learn new features in each layer at the same time [6][7]. The LSTM unit consists of an input gate $i_t$, a forget gate $f_t$, an output gate $o_t$ and a cell state $c_t$. The interaction in a *LSTM* unit is defined as

$$i_t = \sigma(W_i[X_t, H_{t-1}] + b_i), f_t = \sigma(W_f[X_t, H_{t-1}] + b_f)$$
$$g_t = \sigma(W_g[X_t, H_{t-1}] + b_g), o_t = \sigma(W_o[X_t, H_{t-1}] + b_o), \quad (2)$$
$$C_t = f_t \odot C_{t-1} + i_t \odot g_t, \quad H_t = o_t \odot \varepsilon(C_t)$$

where $X_t$ denotes the feature maps obtained by prepositive $t$-stage *Hybrid Block* unit, $C_t$ denotes the cell state that will be fed to the *LSTM* unit of next stage, $H_t$ is the output of current *LSTM* unit and will be sent to the *Convolution layer*, $[\bullet]$ is concatenate operation, $\sigma$ and $\varepsilon$ are the sigmoid and Tanh functions, respectively. In the training step, the input image $I$ will be concatenated with the extracted mask (stage) from the previous stage and then will be fed into the *Hybrid Block*. Initially, $H_{t-1}$ and $C_{t-1}$ are set to 0, which has the same dimension as the output $X$ of the *Hybrid Block* unit, and the mask (0) is set to 0.5 as the initial input. Next, we explain how to constrain the mask extracted from U-ARSE.

**U-ARSE in domains $R$ and $N$.** To extract precise rain streak, we use a constraint to transfer information between the rainy and rain-free domains, and solve the information asymmetry between the two domains. However, since there is no ground truth due to the unsupervised manner of our model, we define two new priors on rain streak information $Att(r)$ and $Att(n)$. The total attention loss of U-ARSE is

$$\mathcal{L}_{att} = \mathcal{L}_{attr} + \mathcal{L}_{attn},$$
$$\mathcal{L}_{attr} = \|Att(r) - \mathbb{N}\|_2^2,$$
$$\mathcal{L}_{attn} = \|Att(n) - \mathbb{Z}\|_2^2,$$

where $\mathbb{N} \sim (0,1)$ is a Gaussian distribution on the interval of $(0, 1)$, $\mathbb{Z}$ is a distribution of all zeros that have same shape as the mask. We set a prior constraint of MSE loss between the rain streak mask $Att(r)$ and the Gaussian distribution $\mathbb{N}$ for the rainy image $r$. Since there is no rain streak in the rain-free image $n$, the distribution of $Att(n)$ must be close to $\mathbb{Z}$ as much as possible. We show some rain streak masks extracted by U-ARSE in Fig.4, where we can see that the learned masks are recurrently improved from stage to stage.

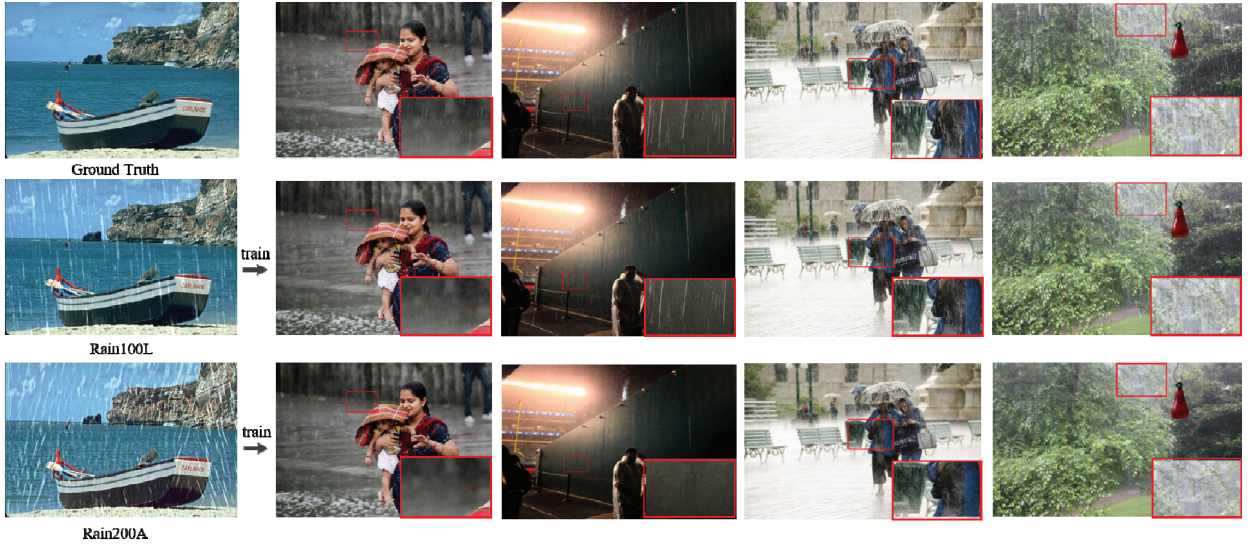

Figure 5. Comparison of real image deraining results using PReNet [24] trained on Rain100L [15] and our Rain200A. The first column is an example of the same rainy image from Rain100L and Rain200A. The first row denotes the real rainy images from SS-TL-Data [27], the second and third rows are the deraining results of PReNet, which are trained on Rain100L and Rain200A respectively. From the results, we can see clearly that the deraining result of PReNet trained on our Rain200A than those of PReNet trained on Rain100L.

### 3.3. Generators and Discriminators

**Generators in domains $R$ and $N$.** As a generator, U-net is commonly-used in the GAN. We also use two generators $G_N$ and $G_R$ that have the same structures as the variant of U-net [28]. The generator includes 16 *Conv-ReLU* blocks. The skip connections are included to prevent generating blurred images. The input of generator is the concatenation of the input image and final attention map from the front U-ARSE. Specifically, $G_N$ uses $r$ and $\mathcal{A}tt(r)$ to generate a rain-free image $n_r$, while $G_R$ uses $n$ and $\mathcal{A}tt(n)$ to generate a rainy image $r_n$. We perform the forward translation as

$$n_r = G_N(\mathcal{A}tt(r), r), \quad r_n = G_R(\mathcal{A}tt(n), n). \quad (4)$$

To be cycle-consistent, we implement the backward pass in the two domains $R$ and $F$. Specifically, U-ARSE is able to extract rain streak information $\mathcal{A}tt(n_r)$ and $\mathcal{A}tt(r_n)$ from the generated samples $n_r$ and $r_n$. Then, $G_R$ can use $n_r$ and $\mathcal{A}tt(n_r)$ to reconstruct the rainy image $\tilde{r}$, while $G_N$ uses $r_n$ and $\mathcal{A}tt(r_n)$ to reconstruct the rain-free image $\tilde{n}$. The backward translation is defined as

$$\tilde{r} = G_R(\mathcal{A}tt(n_r), n_r), \quad \tilde{n} = G_N(\mathcal{A}tt(r_n), r_n). \quad (5)$$

The rainy to rainy branch and the rain-free to rain-free branch in our network can then be constructed based on the forward and backward translations.

**Discriminators in domains $R$ and $N$.** We involve the two adversarial discriminators $D_R$ and $D_N$ for our method, where $D_R$ distinguishes the rainy image $r$ and translated image $r_f$; and similarly $D_N$ aims at distinguishing $n$ and rain-free image $n_r$. The structures of the adversarial discriminators are described in [36], the discriminator uses a multi-scale structure where feature maps at each scale go through three convolutional layers and then are fed into the sigmoid output. To make the generated images look more realistic, we use the adversarial loss in both domains. For the rainy domain $R$, we define the adversarial loss as

$$\begin{aligned} \mathcal{L}_{D_R} = & \; \mathbb{E}_{r \sim p(r)}\left[\log D_R(r)\right] \\ & + \mathbb{E}_{n \sim p(n)}\left[\log\left(1 - D_R\left(G_R(\mathcal{A}tt(n), n)\right)\right)\right], \end{aligned} \quad (6)$$

where $D_R$ maximizes the objective function to distinguish generated rainy images and real rainy images. In contrast, $G_R$ minimizes the loss to make the generated rainy images look similar to real samples in the domain $R$. Similarly, we define the adversarial loss in rain-free domain $F$ as

$$\begin{aligned} \mathcal{L}_{D_N} = & \; \mathbb{E}_{n \sim p(n)}\left[\log D_N(n)\right] \\ & + \mathbb{E}_{r \sim p(r)}\left[\log\left(1 - D_N\left(G_N(\mathcal{A}tt(r), r)\right)\right)\right]. \end{aligned}$$

### 3.4. Objective Function

We describe the objective function of our net for unsupervised training, which is presented as follows:

$$\begin{aligned} \mathcal{L}_{total} = & \; \lambda_{adv}\mathcal{L}_{adv} + \lambda_{att}\mathcal{L}_{att} + \lambda_{cc}\mathcal{L}_{cc} \\ & + \lambda_p\mathcal{L}_p + \lambda_{gmm}\mathcal{L}_{gmm} + \lambda_r\mathcal{L}_r \end{aligned},$$

where the involved losses will be introduced shortly, $\lambda_{adv}$, $\lambda_{att}, \lambda_{cc}, \lambda_p, \lambda_{gmm}$ and $\lambda_r$ are trade-off parameters.

**Constrained two-branch cycle-consistency loss $\mathcal{L}_{cc}$ in domains $R$ and $F$.** For our method, it is better to generate undistinguished rain-free image from $G_N$. But since no paired supervision is provided, the derained image may not keep the color and structure information in images. Inspired by the CycleGAN [1], we use the cycle-consistency loss to ensure the de-rained image $n_r$ to be re-rained to reconstruct the rainy sample and ensure $r_n$ to be translated back into the original rain-free image domain. $\mathcal{L}_{cc}$ can limit the space of generated samples and preserve the contents of the images. We define the loss $\mathcal{L}_{cc}$ in both image domains as

$$\mathcal{L}_{cc} = \mathbb{E}_{r \sim p(r)}\left[\|r - \tilde{r}\|_1\right] + \mathbb{E}_{n \sim p(n)}\left[\|n - \tilde{n}\|_1\right]. \quad (9)$$

**Perceptual loss $\mathcal{L}_p$ in domain $R$.** For the CycleGAN, the generated rain-free samples often contain some unpleasant artifacts. Motivated by [18] that showed that features extracted from pretrained deep networks contain rich semantic information, and their distances can act as the perceptual similarity measure. As such, we utilize the perceptual loss to encode the difference between derained image $n_r$ and the corresponding original rainy image $r$:

$$\mathcal{L}_p = \|\phi_l(n_r) - \phi_l(r)\|_2^2, \quad (10)$$

where $\phi_l(\bullet)$ is the feature extractor of the $l$-th layer of the pretrained CNN. We utilize the $conv_{2,3}$ layer of VGG-16 network [5] pre-trained on ImageNet [37].

**GMM loss $\mathcal{L}_{gmm}$ in domain $R$.** This loss describes the rain streaks from input rainy images by a GMM:

$$S \sim \sum_{k=1}^{K} \pi_k \mathcal{N}(S \mid \mu_k, \Sigma_k), \quad (11)$$

where $S$ is rain streak, $K$ is the number of mixture components, $\pi_k, \mu_k$ and $\Sigma_k$ are mixture coefficients, Gaussian distribution means and variance, respectively. The negative log likelihood function of rain streak $S$ is defined as $\mathcal{L}_{gmm}$:

$$\mathcal{L}_{gmm}(S; \Pi, \Sigma) = -\sum_{k=1}^{K} \log \sum_{k=1}^{K} \pi_k \mathcal{N}(S_n \mid 0, \Sigma_k), \quad (12)$$

where $S_n = r - n_r$ denotes the rain streak to be learned and extracted from the input rainy image, which is equivalent to $R_n$, $\Pi = \pi_1, ..., \pi_K$, $\Sigma = \Sigma_1, ..., \Sigma_K$ and $N$ is the number of samples. Note that the intractable loss in Eq. (10) can be iteratively solved by the Expectation Maximization (EM) method [27][38], and details can be referred to [27].

**Reconstructive loss $\mathcal{L}_r$ in domain $R$.** We introduce the reconstruct loss to encode the mismatch between the rainy image $r$ and the recovered rainy image $r'$. We minimize the discrepancy between composite image $r' = \mathcal{A}tt(r) + n_r$ and the original rainy image $r$, defined as follows:

$$\mathcal{L}_r = \|r' - r\|_2^2. \quad (13)$$

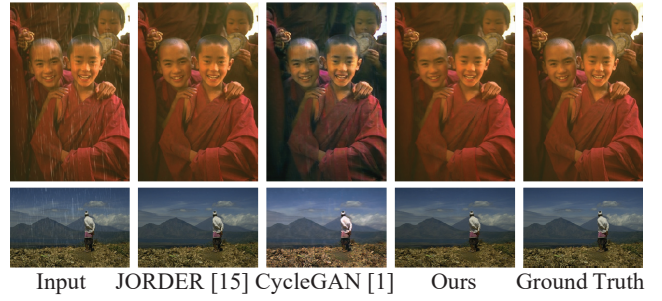

Input    JORDER [15]    CycleGAN [1]    Ours    Ground Truth

Figure 6. Comparison with other deraining methods on Rain100L and Rain12, where JORDER [15] and CycleGAN [1] are supervised and unsupervised methods, respectively.

### 3.5. Testing Procedure by our Network

We describe the testing process using our net. For testing, the main purpose is to obtain the rain-free images from the first branch, thus the branch $r_n = G_R(\mathcal{A}tt(n), n)$ is removed. Given a test rainy image $r$, we first use U-ARSE to extract rain streak information $\mathcal{A}tt(r)$. Then, $G_N$ uses the extracted rain streak information $\mathcal{A}tt(r)$ and original test rainy image $r$ to generate the derained image $n_r$ of $r$ as follows:

$$n_r = G_N(\mathcal{A}tt(r), r). \quad (14)$$

### 3.6. Created Rain200A Dataset

For SID, most existing datasets are synthetic rainy images, but the synthetic rain streaks are well-designed, so they are not as real as real rain streaks. As a result, the deep nets trained on the synthetic rainy images usually produce unsatisfactory results when handling real images. However, real-world rainy images and their ground truths are usually difficult to collect, and most of them have lower resolution, which may cause the converge issue in the training process. Besides, without the ground truth, the datasets are also not suitable for the supervised deraining methods.

In this paper, we introduce a new rainy image dataset named Rain200A, which is automatically generated by our net on Rain100L. Specifically, we use 200 rain-free images and sent them to the second branch $n \to r_n \to \tilde{n}$ of our net, then our net will automatically add rain streaks into each rain-free image $n$ and generate rainy image $r_n$ as

$$r_n = G_R(\mathcal{A}tt(n), n). \quad (15)$$

The advantage of Rain200A is by automatically adding rain streak, so it makes the rain streaks have more various shapes and directions which will be adapted to the real rain streaks than original Rain100L that is manually created by Photoshop. We use the supervised PReNet [24] to show the effectiveness of the generated irregular rain streaks in our Rain200A experimentally. Note that the compare result is fair due to two reasons: (1) we perform PReNet under the same training and testing settings; (2) both Rain100L and Rain200A have the same number of image pairs.

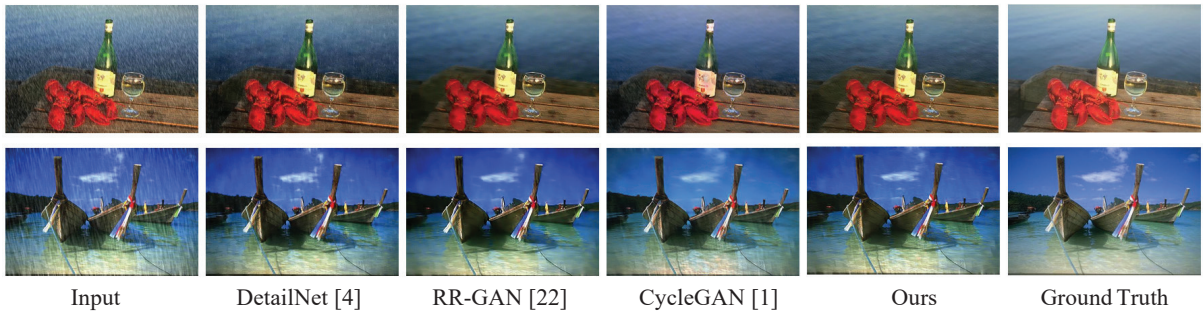

| Input | DetailNet [4] | RR-GAN [22] | CycleGAN [1] | Ours | Ground Truth |

Figure 7. Comparison with other state-of-the-arts on Rain800, where RR-GAN [22] and CycleGAN [1] are unsupervised nets.

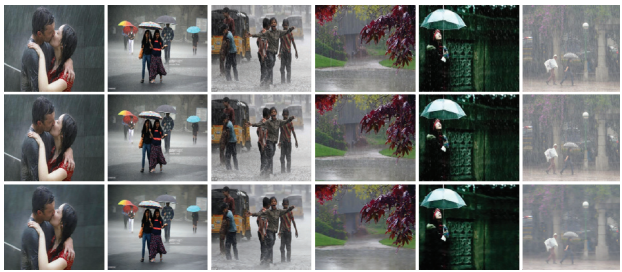

Figure 8. Comparison with the supervised PReNet [24] on several real-word images, where we show the original rainy images (first row), the derained images of PReNet [24] (second row) and the derained images of our network (third row), respectively.

The comparison results are shown in Fig.5, where we see that the deraining result of PReNet trained on Rain200A is better than that of PReNet trained on Rain100L, even for the rain streaks on white background. Thus, our Rain200A can be potentially used to train a net to solve the real image removal task. Note that our network can also be used to build more complex non-artificial datasets to solve the SID task of real images, which will be explored in future.

## 4. Experiments and Results

We evaluate the performance of each algorithm for SID, and the results are compared with related deep frameworks. The network is trained using the Pytorch framework [19] in Python environment on a NVIDIA GeForce GTX 1080i GPU with 12GB memory. For training, a 216×216 image is randomly cropped from input image (or its horizontal flip) of original size. Adam [20] is employed as the optimization algorithm with a mini-batch size of 1. We use a weight decay of 0.0001 and a momentum of 0.9. The models are trained for total 400 epochs. The learning rate starts from 0.0001 and decry with a policy of Pytorch after 200 epochs. We set the parameters $\lambda_{adv}=1$, $\lambda_{att}=10$, $\lambda_{cc}=10$, $\lambda_p=0.01$, $\lambda_{gmm}=10$ and $\lambda_r=10$ for our network. All parameters are defined via cross validation using the validation set, and the whole network is trained in unsupervised mode.

### 4.1. Dataset and Evaluation Metrics

**Datasets.** We evaluate each method using the synthetic and real-world datasets. The synthetic dataset includes: (1) **Rain100L** [15], where 200 image pairs are used for training and 100 image pairs are for testing; (2) **Rain800** [13] has 700 synthesized images for training and 100 images for testing; (3) **Rain12** [10]: 12 rainy and clean image pairs. Since Rain12 has few samples, we directly use the trained model on Rain100L. Real datasets are: (1) **SS-TL-Data** [27] has 147 images without the ground truth; (2) **SPANet-Data** [25] that contains 1000 images with the ground truth.

**Evaluation metrics.** For images with ground truth, we evaluate each method by two commonly used metrics, i.e., Peak Signal-to-Noise Ratio (PSNR) [21] and Structural Similarity Index (SSIM) [17]. For the cases without ground truth, i.e., SS-TL-Data, we only provide visual results.

**Compared methods.** The deraining result of our net is compared with those of two model-driven algorithms (i.e., DSC [11] and GMM [10]), four supervised deep nets (i.e., DetailNet [4], JORDER [15], RESCAN [12], and PReNet [24]), one semi-supervised deep net SS-TL [27], and two unsupervised deep nets (i.e., CycleGAN [1] and RR-GAN [22]). Since our net is unsupervised method, CycleGAN [1] and RR-GAN [22] will be mainly compared with, although our net obtains highly-competitive and even better results than the related semi-supervised/supervised deep nets.

### 4.2. Deraining Results on Synthetic Rainy Images

We first evaluate each method on three popular synthetic rain image datasets, i.e., Rain100L [15], Rain12 [10] and Rain800 [4] in Table 1. For the fair comparison, we use the same settings for training or directly use the code provided by authors. For some methods, we directly use the results proposed in [32]. We find that: (1) compare with unsupervised methods, our net achieves better performance than CycleGAN [1], especially on Rain100L and Rain12. Besides, the performance of our net on Rain800 is better than CycleGAN [1] and RR-GAN [22]; (2) the performance of the other methods declined fast from Rain100L to Rain800, while our net can remain stable. (3) The semi-supervised SS-TL [27] should have obtained better results than both supervised and unsupervised nets on synthesis datasets, but its results are even worse than our unsupervised net.

Table 1. Comparison with the different types of methods using the PSNR and SSIM on four datasets. Since RR-GAN has no available code in practice, we only compare with the result from its original paper using the same metrics on the Rain800 dataset.

| Datasets | | Rain100L | | Rain12 | | Rain800 | | SPANet-Data | |
|---|---|---|---|---|---|---|---|---|---|
| Metrics | | PSNR | SSIM | PSNR | SSIM | PSNR | SSIM | PSNR | SSIM |
| Model-based methods | DSC [11] | 27.34 | 0.849 | 30.07 | 0.866 | 18.56 | 0.600 | 34.95 | 0.942 |
| | GMM [10] | 29.05 | 0.872 | 32.14 | 0.916 | 20.46 | 0.730 | 34.30 | 0.943 |
| Supervised methods | DetailNet [4] | 32.38 | 0.926 | 34.04 | 0.933 | 21.16 | 0.732 | 34.70 | 0.934 |
| | JORDER [15] | 36.61 | 0.974 | 33.92 | 0.953 | 22.24 | 0.776 | / | / |
| | RESCAN [12] | 38.52 | 0.981 | 36.43 | 0.952 | 24.09 | 0.841 | 34.70 | 0.938 |
| | PReNet [24] | 37.45 | 0.979 | 36.66 | 0.961 | 26.97 | 0.898 | 35.06 | 0.944 |
| Semi-supervised method | SS-TL [27] | 32.37 | 0.926 | 34.02 | 0.935 | / | / | 34.85 | 0.936 |
| Unsupervised methods | CycleGAN [1] | 24.61 | 0.834 | 21.56 | 0.845 | 23.95 | 0.819 | 22.40 | 0.860 |
| | RR-GAN [22] | / | / | / | / | 23.51 | 0.757 | / | / |
| | Ours | 31.49 | 0.936 | 34.44 | 0.952 | 24.32 | 0.842 | 34.12 | 0.950 |

Table 2. Deraining results with different losses on Rain100L.

| | $\mathcal{L}_{adv} + \mathcal{L}_{cc}$ | w. | w. | w. | $\mathcal{L}_{total}$ |
|---|---|---|---|---|---|
| PSNR | 28.59 | 30.51 | 30.83 | 31.12 | 31.49 |
| SSIM | 0.902 | 0.921 | 0.928 | 0.931 | 0.936 |

We also visualize derained images on the three synthesis datasets in Fig.6 and Fig.7. We find that our net performs better than CycleGAN obviously with the enhanced results than the supervised DetailNet method on Rain800, which keeps consistent with the numerical results in Table 1.

### 4.3. Deraining Results on Real Rainy Images

We also evaluate all algorithms on two real rainy image datasets, namely, SS-TL-Data [27] and SPANet-Data [25]. Since SPANet-Data has the corresponding ground truth, so it can be evaluated using numerical metrics. In this study, all methods are trained on the Rain100L dataset. From the results in Table 1, our network obtains competitive PSNR values and obtains the best results on SSIM metric compared with other related nets on SPANet-Data.

For SS-TL-Data, we visualize some deraining results in Fig.8. Since our network is trained in unsupervised mode, we mix the training set of Rain100L and 69 real images to train a model, while PReNet [24] is a supervised method that is trained on original training set of Rain100L. From Fig.8, we see that our network performs much better than PReNet [24], while PReNet [24] leaves many rain streaks.

### 4.4. Ablation Study

We mainly discuss the selection of different loss function in our net. During training process, although the total loss $\mathcal{L}_{total}$ can prompt the weight converge of our network and obtain better results, it is necessary to explore which loss plays a more important role. The numerical results with different loss functions on Rain100L are shown in Table 2, where $\mathcal{L}_{adv} + \mathcal{L}_{cc}$ is the basic loss function of CycleGAN, w. $\mathcal{L}_p$ corresponds to the added perceptual loss to basic loss function, and other columns are defined by the same rules, and the final $\mathcal{L}_{total}$ means the loss function used in our net. We can find that our net achieves the best performance, i.e., the losses are all important and useful for our net.

## 5. Conclusion and Future Work

We proposed a new attention-guided deraining network by a constrained CycleGAN. Compared with existing rain removal methods that attempt to use paired information for training, we present a novel rain streak extractor U-ARSE to extract the rain streak masks stage by stage in unsupervised manner. We also design two constrained branches for our network, which use the rainy image/rain-free image to generate rain-free image/rainy image and then reconstruct the rainy image/rain-free image by generators. In addition, we construct a new rainy dataset Rain200A, which can help the supervised model to work better on real rainy images.

We evaluated our net on synthetic and real rainy images. The obtained results show that our network is superior to most existing unsupervised deraining networks, and is also competitive to related supervised networks. In future, we shall consider extending our method to the semi-supervised mode to enable it to make use of existing synthesis paired data and real rainy images jointly. We will also continue studying the automatic generation of rainy image datasets, especially on generating larger-scale rainy image datasets (that are very lacking in the area to date) with more diverse irregular rain streaks to help existing supervised SID nets to obtain better performance on real-world rainy images.

## Acknowledgement

This work is partially supported by ##################
#####################################.

# References

[1] J. Zhu, T. Park, P. Isola, A. Efros. Unpaired image-to-image translation using cycle-consistent adversarial networks. In *ICCV*, 2017.

[2] S. Ren, K. He, R. Girshick, and J. Sun. Faster R-CNN: Towards Real-Time Object Detection with Region Proposal Networks. *IEEE TPAMI*, vol. 39, no. 6, pp. 1137-1149, 2017.

[3] X. Fu, J. Huang, X. Ding, Y. Liao, J. Paisley. Clearing the Skies: A Deep Network Architecture for Single-Image Rain Removal. *IEEE TIP*, vol. 26, no. 6, pp. 2944-2956, 2017.

[4] X. Fu, J. Huang, D. Zeng, Y. Huang, X. Ding, and J. Paisley. Removing Rain from Single Images via a Deep Detail Network. In *CVPR*, pp.1715-1723, 2017.

[5] K. Simonyan and A. Zisserman. Very deep convolutional networks for large-scale image recognition. In *ICLR*, 2015.

[6] K. He, X. Zhang, S. Ren, and J. Sun. Deep Residual Learning for Image Recognition. In *CVPR*, 2016.

[7] G. Huang, Z. Liu, L. Maaten, and K. Weinberger. Densely Connected Convolutional Networks. In *CVPR*, 2017.

[8] E. David, R. Jason, F. Rob, and L. Yann. Understanding deep architectures using a recursive convolutional network. In *ICLR*, 2014.

[9] G. Li, Y. Xie, L. Lin, Y. Yu. Instance-Level Salient Object Segmentation. In *CVPR*, pp. 247-256, 2017.

[10] Y. Li, R. T. Tan, X. Guo, J. Lu, and M. S. Brown. Rain Streak Removal Using Layer Priors. In *CVPR*, 2016.

[11] Y. Luo, Y. Xu, and H. Ji. Removing Rain from a Single Image via Discriminative Sparse Coding. In *ICCV*, 2015.

[12] X. Li, J. Wu, Z. Lin, H. Liu, and H. Zha. Recurrent squeezeand-excitation context aggregation net for single image deraining. In *ICCV*, pp. 262-277, 2018.

[13] H. Zhang, V. Sindagi, and V. Patel. Image de-raining using a conditional generative adversarial network. In *CVPR*, 2017.

[14] H. Zhang and V. M. Patel. Density-Aware Single Image De-raining Using a Multi-stream Dense Network. In *CVPR*, pp.695-704, 2018.

[15] W. Yang, R. T. Tan, J. Feng, J. Liu, Z. Guo, and S. Yan. Deep Joint Rain Detection and Removal from a Single Image. In *CVPR*, pp. 1685-1694, 2017.

[16] W. Yang, R. Tan, J. Feng, J. Liu, S. Yan, and Z. Guo. Joint rain detection and removal from a single image with contextualized deep networks. *IEEE Trans. Pattern Analysis and Machine Intelligence*, vol. PP, no. 99, 2019.

[17] Z. Wang, A. C. Bovik, H. R. Sheikh, and E. P. Simoncelli. Image quality assessment: from error visibility to structural similarity. *IEEE TIP*, vol. 13, no. 4, pp. 600-612, 2004.

[18] J. Johnson, A. Alahi, and F. L. Perceptual losses for real-time style transfer and super-resolution. *arXiv preprint arXiv:* 1603.08155, 2016.

[19] A. Paszke, S. Gross, S. Chintala, G. Chanan, E. Yang, Z. DeVito, Z. Lin, A. Desmaison, L. Antiga, A. Lerer. Automatic differentiation in pytorch. In *NIPS Workshop*, 2017.

[20] D. P. Kingma and J. Ba. Adam: A method for stochastic optimization. In *ICLR*, San Diego, USA, 2015.

[21] Q. Huynh-Thu and M. Ghanbari. Scope of validity of PSNR in image/ video quality assessment. *Electronics Letters*, vol. 44, no. 13, pp. 800-801, 2008.

[22] H. Zhu, X Peng, Joey T. Zhou, S. Yang, V. Chandrasekhar, L. Li, and J. Lim. RR-GAN: Single Image Rain Removal Without Paired Information. In *AAAI*, 2019.

[23] Z. Tang, W. Jiang, Z. Zhang, M. Zhao, L. Zhang, M. Wang. DenseNet with Up-Sampling Block for Recognizing Texts in Images. *Neural Computing and Applications*, 2019.

[24] D. Ren, W. Zuo, Q. Hu, P. Zhu, and D. Meng. Progressive image deraining networks: a better and simpler baseline. In *CVPR*, 2019.

[25] T. Wang, X. Yang, K. Xu, S. Chen, Q. Zhang, and R. W. H. Lau. Spatial attentive single-image deraining with a high quality real rain dataset. In *CVPR*, 2019.

[26] Y. Wei, Z. Zhang, H. Zhang, R. Hong, and M. Wang. A Coarse-to-Fine Multi-stream Hybrid Deraining Network for Single Image Deraining. In *ICDM*, 2019.

[27] W. Wei, D. Meng, Q. Zhao, Z. Xu, Y. Wu. Semi-supervised transfer learning for image rain removal. In *CVPR*, pp. 3877-3886, 2019.

[28] R. Qian, R. Tan, W. Yang, J. Su, J. Liu. Attentive generative adversarial network for raindrop removal from a single image. In *CVPR*, 2018.

[29] I. Goodfellow, J. Pouget-Abadie, M. Mirza, B. Xu, D. Warde-Farley, S. Ozair, A. Courville, Y. Bengio. Generative adversarial nets. In *NeurIPS*, pp. 2672–2680, 2014.

[30] D. Martin, C. Fowlkes, D. Tal, and J. Malik. A database of human segmented natural images and its application to evaluating segmentation algorithms and measuring ecological statistics. In *ICCV*, pp. 416-423, 2001.

[31] B. Lu, J. Chen, and R. Chellappa. Unsupervised Domain-Specific Deblurring via Disentangled Representations. In *CVPR*, 2019.

[32] H. Wang, M. Li, Y. Wu, Q. Zhao, and D. Meng. A Survey on Rain Removal from Video and Single Image. *arXiv preprint arXiv:*1909.08326, 2019.

[33] Y. Yuan, S. Liu, J. Zhang, Y. Zhang, C. Dong, and L. Lin. Unsupervised image super-resolution using cycle-in-cycle generative adversarial networks. In *CVPR Workshops*, pages 701–710, 2018.

[34] D. Engin, A. Genç, and H. Ekenel. Cycle-Dehaze: Enhanced CycleGAN for Single Image Dehazing. In *CVPR Workshops : NTIRE*, 2018.

[35] X. Shi, Z. Chen, H. Wang, D. Yeung, W. Wong, and W. Woo. Convolutional lstm network: A machine learning approach for precipitation nowcasting. In *NeurIPS*, 2015.

[36] H. Lee, H. Tseng, J. Huang, M. Singh, and M. Yang. Diverse Image-to-Image Translation via Disentangled Representations. In *ECCV*, 2018.

[37] J. Deng, W. Dong, R. Socher, L. Li, K. Li, F. Li. ImageNet: A Large-Scale Hierarchical Image Database. In *CVPR*, 2009.

[38] A. Dempster, N. Laird, and D. Rubin. Maximum likelihood from incomplete data via the em algorithm. *Journal of the Royal Statistical Society*, 39(1), pages 1–22, 1977.

[39] L. Kang, C. Lin, and Y. Fu. Automatic single-image-based rain streaks removal via image decomposition. *IEEE Trans. on Image Processing*, 21(4):1742–1755, 2012.

[40] H. Zhang, V. Sindagi, and V. M. Patel. Image de-raining using a conditional generative adversarial network. In *CVPR*, 2017.